\documentclass[10pt,twocolumn,letterpaper]{article}

\usepackage{cvpr}
\usepackage{times}
\usepackage{epsfig}
\usepackage{soul}
\usepackage{graphicx}
\usepackage{amsmath}
\usepackage{amssymb}
\usepackage{graphicx}
\usepackage{amsmath,amssymb} 
\usepackage{color}
\usepackage{enumitem}
\usepackage{rotating}
\usepackage{times}
\usepackage{epsfig}
\usepackage{graphicx}
\usepackage{amsmath}
\usepackage{amssymb}
\usepackage{amsfonts}
\usepackage{empheq}
\usepackage{framed, color}
\usepackage{xcolor}
\usepackage{graphics}
\usepackage{graphicx}
\usepackage[]{graphicx} 
\usepackage{epsfig} 
\usepackage{subfigure}
\usepackage{algorithm}
\usepackage{algorithmic}
\usepackage{stmaryrd}
\usepackage[mathscr]{eucal}
\usepackage{lineno}
\usepackage{color}
\usepackage{filecontents}
\usepackage{subfigure}
\usepackage{multirow}

\usepackage{filecontents}
\usepackage{verbatim} 
\usepackage{tabularx}
\usepackage{lineno}
\usepackage{setspace}
\usepackage{multirow}
\usepackage{textcomp,booktabs}
\usepackage{caption}
\usepackage{makecell}
\usepackage{authblk}

\usepackage{array, boldline, rotating}

\newcommand{\thickhline}{\hlineB{4}}

\def\x{{\bf x}}

\def\w{{\bf w}}
\def\0{{\bf 0}}
\def\1{{\bf 1}}

\graphicspath{{./figures/}}   

\newcommand*{\colorboxed}{}
\def\colorboxed#1#{%
	  \colorboxedAux{#1}%
}

\newcommand*{\colorboxedAux}[3]{%
	\begingroup
	\colorlet{cb@saved}{.}%
	\color#1{#2}%
	\boxed{%
		\color{cb@saved}%
		#3%
	}%
	\endgroup
}

\makeatletter
\renewcommand\subsubsection{\@startsection{subsubsection}{3}{\z@}%
                                     {-3.25ex\@plus -1ex \@minus -.2ex}%
                                     {-1.5ex \@plus -.2ex}
                                     {\normalfont\normalsize\bfseries}}
\makeatother

\cvprfinalcopy 

\def\cvprPaperID{3007} 

\ifcvprfinal\fi
\begin{document}

\title{Modularized Textual Grounding for Counterfactual Resilience}

\author{Zhiyuan Fang$^{1}$, Shu Kong$^{2}$, Charless Fowlkes$^{2}$, Yezhou Yang$^{1}$\\
    \ $^{1}${\tt \{zy.fang, yz.yang\}@asu.edu} \ \ \ \ \ \ \ \ \ \ \ Arizona State University, Tempe, USA\\  
    \ $^{2}${\tt \{skong2, fowlkes\}@ics.uci.edu} \ \ \ \ \ \ \ \ University of California, Irvine, USA 
    \vspace{-5mm} \\}
\maketitle

\begin{abstract}
Computer Vision applications often require a textual grounding module with precision, interpretability, and resilience to counterfactual inputs/queries. To achieve high grounding precision, current textual grounding methods heavily rely on large-scale training data with manual annotations at the pixel level. Such annotations are expensive to obtain and thus severely narrow the model's scope of real-world applications. Moreover, most of these methods sacrifice interpretability, generalizability, and they neglect the importance of being resilient to counterfactual inputs. To address these issues, we propose a visual grounding system which is 1) end-to-end trainable in a weakly supervised fashion with only image-level annotations, and 2) counterfactually resilient owing to the modular design. Specifically, we decompose textual descriptions into three levels: entity, semantic attribute,  color information, and perform compositional grounding progressively. We validate our model through a series of experiments and demonstrate its improvement over the state-of-the-art methods. In particular, our model's performance not only surpasses other weakly/un-supervised methods and even approaches the strongly supervised ones, but also is interpretable for decision making and performs much better in face of counterfactual classes than all the others.
\vspace{-4mm}

\end{abstract}

\section{introduction}

\begin{figure}[t]
\begin{center}
\includegraphics[width=1.0\linewidth]{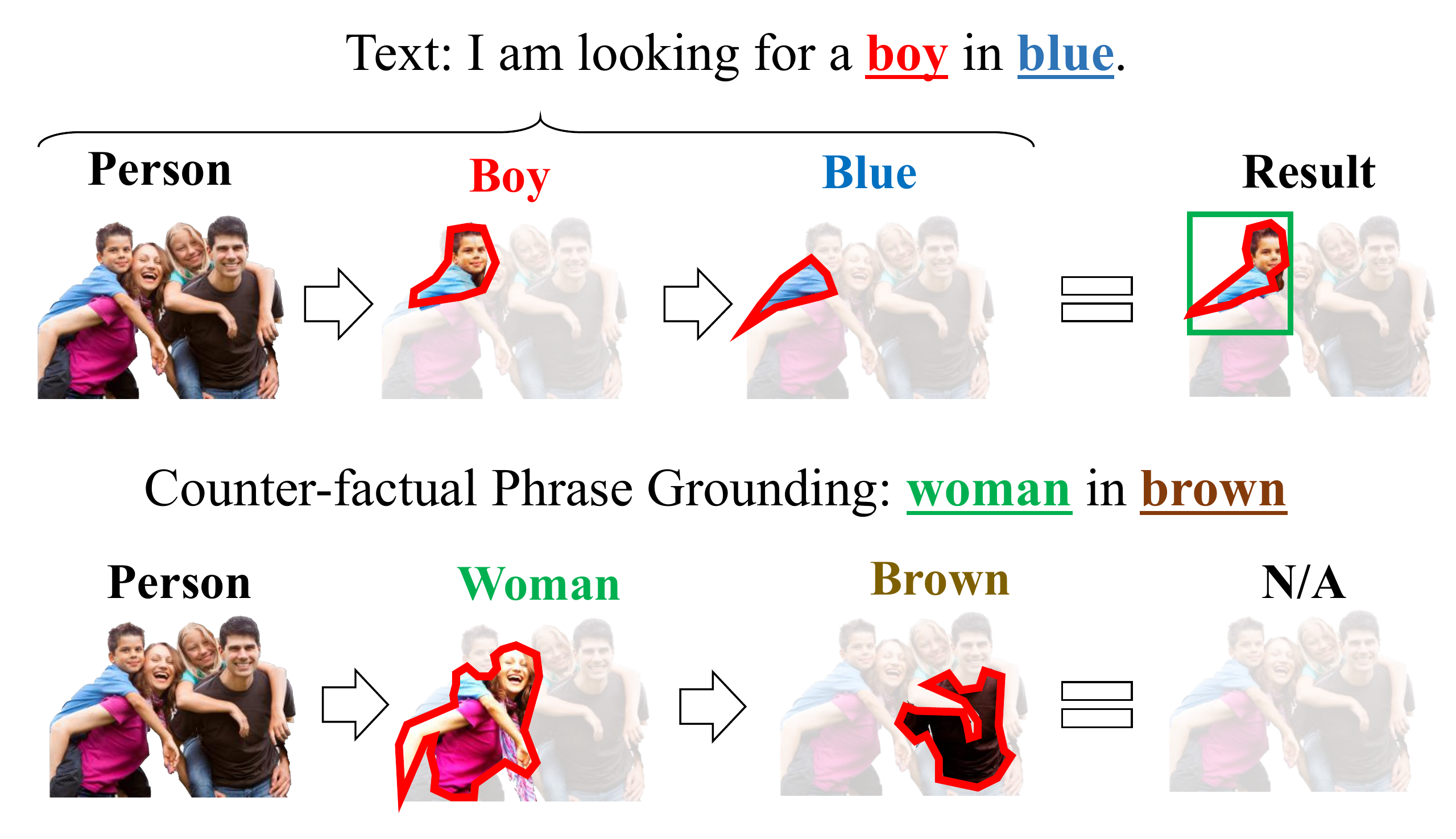}
\end{center}
\vspace{-2mm}
\caption{Illustration of our textual grounding framework that decomposes textual descriptions
into three levels: entity, semantic attributes and color information.
As an example, for textual grounding from the sentence shown above,
our system localizes the entity (person), semantic attributes (boy, woman), the color blue,
and progressively produces the final textual grounding by combining results.
Note that owing to the decomposable description and modular design,
our system is highly interpretable and resilient to counterfactual inputs/qeueries (bottom row).}
\vspace{-2mm}
\label{fig:abstract}
\end{figure}

\begin{figure*}[t]
\begin{center}
\includegraphics[width=1.0\linewidth]{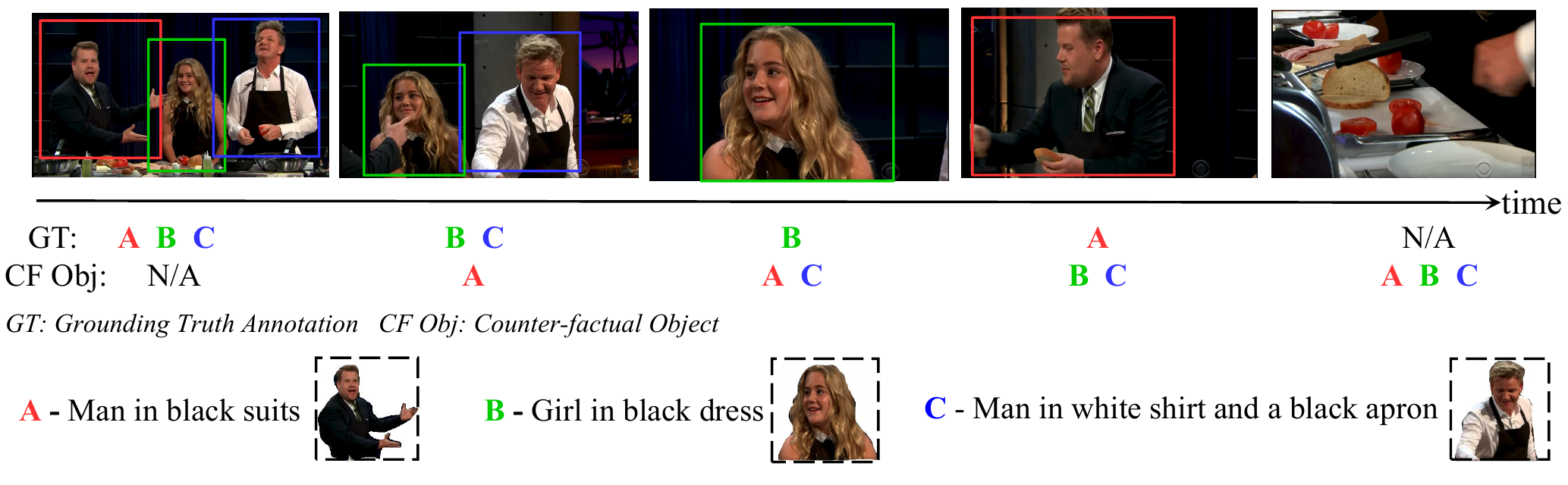}
\end{center}
\vspace{-6mm}
\caption{
Examples of counterfactual objects and applying our system to video captioning alignment.
Although there are three persons in the beginning of the video,
they may disappear later for some frames.
This poses a challenge for video captioning,
our system acts as a tool to ground the object temporally and correct mismatched description and frames.}
\vspace{-4mm}
\label{fig:ca_demo}
\end{figure*}

Deep neural networks have spawned a flurry of successful work on
various computer vision applications,
from modular tasks like object instance detection~\cite{hariharan2015hypercolumns,he2017mask,kong2018recurrent, zhang2017range, wang2017multi}
and semantic segmentation~\cite{long2015fully,chen2018deeplab},
to more complex multi-modal ones like visual question answering (VQA)~\cite{antol2015vqa}
and image captioning~\cite{bengio2015scheduled,johnson2016densecap}.
For complex vision applications (e.g., visual search engine and video auto-captioning),
it is critical to build a reliable textual grounding system, which connects natural language descriptions and image regions~\cite{yu2018mattnet, kazemzadeh2014referitgame, huang2018finding, rohrbach2016grounding, yeh2018unsupervised}.

Current methods typically formulate the textual grounding problem as a search process or image-text matching.
For example, \cite{rohrbach2016grounding} proposed textual-visual feature matching by reconstruction loss.
\cite{chen2017query} fulfills textual grounding with two steps:
the generation of object proposals and  match with the query. \cite{yu2018mattnet} utilizes pre-trained module to conduct searching and matching progressively. Given a novel image and queries, these models return the proposals which yield the highest matching score/probability as the final output.
Although they achieve state-of-the-art performance in terms of grounding precision, they rely on a large-scale training sets with manually annotated bounding boxes on the objects of interest.
This inevitably prevents them from generalizing to other data domains which have no such fine-grained manual annotations for model training or fine-tuning~\cite{yeh2018unsupervised}.

Moreover, these models lack the interpretability for decision making and the resilience to counterfactual queries,
which often appear jointly to make these models even more vulnerable in real-world applications~\cite{hendricks2018generating, wachter2017counterfactual, dhurandhar2018explanations, doshi2017accountability}.
For example,
as demonstrated by Figure~\ref{fig:abstract},
if one is asking ``who is the woman in blue shirt in the image'',
a good model should return nothing instead of the closest person or someone with high matching score.
Even more preferred,
the model should explain why the decision is made in addition to the final grounding result.
The interpretability and counterfactual resilience properties are also useful in literature and
practical deployment.
As demonstrated by another example about our application to correcting video auto-captioning, as shown in Figure~\ref{fig:ca_demo} (details in Section \ref{sec:app}).
There exist three people in the first frame, while they may disappear in the following frames
but the captioning are still not updated.
Our counterfactually resilient grounding system is able to correct captioning mis-alignment issue.

In this work,
we propose to modularize the textual grounding system by decomposing the textual description into multiple components,
and perform grounding progressively through these components towards the final output.
Recently, modular design is being advocated in the
community~\cite{hu2017modeling, hu2018explainable,yu2018mattnet},
mainly focusing on visual-question-answering and referring expression visual matching.
We show that such a modular design also increases the interpretability of our textual grounding system, that it explains along the way how the final decision is being made.
It is worth noting that the modular design supports diverse training protocols to learn each component.
Therefore, to alleviate the requirement for large-scale fine-grained manual annotations (e.g., bounding box),
we propose to train our entity grounding module in a weakly supervised manner which only needs
image level labels. We note that such data are easy to obtain, e.g., from internet search engine or social media
with image tags \cite{hartmann2012weakly, bergamo2010exploiting, chen2014event}.

To validate our system,
we carry out extensive experiments on the COCO dataset~\cite{lin2014microsoft} and
Flickr30k Entities dataset~\cite{plummer2015flickr30k}.
We show that our system outperforms other weakly-supervised methods on textual grounding and even surpasses some
strongly-supervised approaches.
By introducing another dataset consisting of counterfactual cases,
we emphasize that our system performs remarkably better than other methods w.r.t counterfactual resilience. To summarize our contributions:
\begin{enumerate}[noitemsep,nolistsep]
  \item We propose a textual grounding system with modular design. Together with the decomposition of textual descriptions, it allows for more diverse and specialized training protocols for each components.
  \item We collect a counterfacutal textual grounding test set, and show that our system achieves better interpretability and resilience to counterfactual testing.
  \item We demonstrate practical applications based on our system and expect future explorations based on our work.
\end{enumerate}
In the rest of the paper,
we first review related work,
then describe our system in Section~\ref{sec:system}.
We elaborate our training procedure and demonstrate the effectiveness of our system through experiment in
Section~\ref{sec:exp} and broad application in Section~\ref{sec:app}, respectively,
before concluding in Section~\ref{sec:conclusion}.

\section{Related Work}
Multi-modal tasks, eg. assistive visual search~\cite{cai2004hierarchical, la1998combining}  and image captioning~\cite{you2016image, vinyals2017show}, has been studied for decades in the community. While those tasks are classical topics in computer vision and natural language processing,
current advancement has further energized it by interplaying
vision (images) and language (high-level guide) for practical applications. Specific examples include referring expressing understanding~\cite{nagaraja2016modeling,hu2017modeling} and reasoning-aware visual-question-answering~\cite{hu2017learning}.

State-of-the-art textual grounding methods \cite{yu2018mattnet, hu2016natural, rohrbach2016grounding, plummer2015flickr30k, yeh2017interpretable, luo2017comprehension, fang2018weakly} are based on deep neural networks and relying on
large-scale training data with manual annotations for the object bounding box and relationship between
phrases and figures/objects.
This setup largely limits their broad applications as such strong supervision is expensive to obtain,
and they also lack interpretability and resilience to counterfactual cases which do not appear in training.

Weakly supervised learning receives increasing attention~\cite{deselaers2010localizing,
pandey2011scene, cinbis2017weakly, mahajan2018exploring,pathak2015constrained,pinheiro2015image,xu2014tell}.
It focuses on learning granular detectors given only coarse annotations.
This is of practical significance as granular annotations (e.g., bounding boxes and pixel-level labels) are
much more expensive to obtain compared to coarse image-level annotations.
Recent study shows that weakly supervised methods can even outperform the strongly supervised method for image classification~\cite{mahajan2018exploring}.
Unlike current work,
we perform weakly-supervised learning for textual grounding, including training for both entity grounding
and textual-visual matching through a progressive modular procedure.

Modular design is also receiving more attention recently, mainly for
complex systems like visual-question-answering or image captioning~\cite{hu2017modeling, hu2018explainable,
yu2018mattnet}.
Such modular design is carried out by realizing some linguistic structures.
In our work,
we propose to decompose the query textual description into progressive levels,
each of which is passed to a corresponding module,
and then produce the final grounding result by progressively merging the intermediate results.
In this way,
our system enjoys high interpretability and resilience to counterfactual inputs.

\section{Modularized Textual Grounding System}
\label{sec:system}
To obtain better interpretability and counterfactual resilience,
we propose to modularize the our whole textual grounding system
by decomposing the textual descriptions into multiple levels,
each of which is passed to a specific module to process.
We generate the final grounding result by progressively merging intermediate
results from these modules.

Without losing generalization,
in this work,
we decompose the textual descriptions into three levels,
and progressively process them with three different modules, respectively:
entity grounding module $M_e$,
semantic attribute grounding module $M_a$, and color grounding module $M_c$.
We extracted phrases/words that belong to such three levels from text, and feed them into their corresponding sub-modules.
We note that such a modular design allows for training different modules using different specialized protocols,
e.g., fully supervised learning or weakly supervised learning,
while also enables end-to-end training.
For the final grounding heat map $G$,
we merge progressively the intermediate results from these modules as below (see Figure \ref{fig:pipeline}):
\begin{equation}
\begin{split}
G & = M_e \cdot (M_a + M_c).
\label{eq:att}
\end{split}
\end{equation}

In practice, we observe that such a merging approach achieves the best
performance, better than the straightforward multiplicative or additive fusion.
This is because that the entity grounding sets the object constraints, and the summation
over the attribute and color modules interpretably delivers how the final results are generated,
though they may partially cover some regions belonging to the object of interest.
In the remaining of this section,
we elaborate each of the three modules and the adopted training protocols.

\subsection{Entity Grounding Module (M$_{e}$)}

\begin{figure*}[t]
\begin{center}
\includegraphics[width=1.0\linewidth]{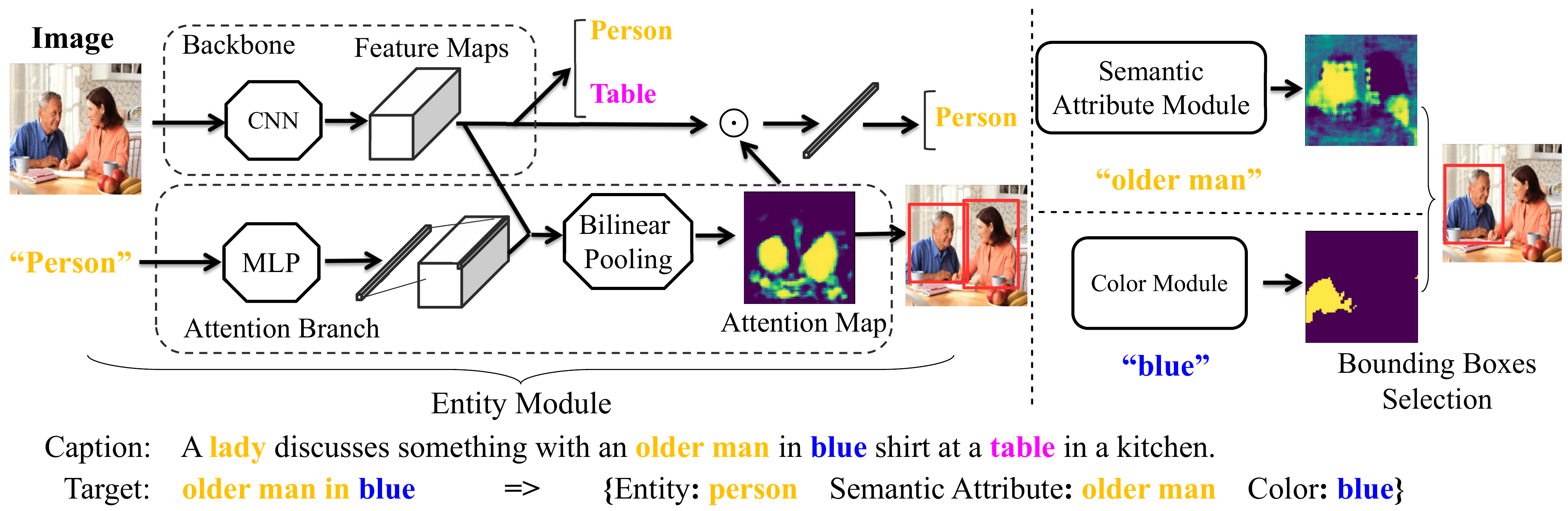}
\end{center}
\caption{Illustrative diagram for our entity grounding module (left) and the whole
textual grounding system (right). The textual phrase is first decomposed into sub-elements, e.g., ``older man in blue" can be parsed to ``person'' category with ``older man'' and ``blue'' to be it's attributes, and later fed into corresponding sub-module. The bounding boxes are generated and selected based upon the merged attention maps. We train the entity/semantic attribute grounding module in a weakly supervised fashion with a attention mechanism. The semantic attribute module also adopt similar architecture of entity module, however with a dictionary learning loss. (best viewed in color)}
\vspace{-2mm}
\label{fig:pipeline}
\end{figure*}

To overcome the limitation of current methods that require expensive manual annotations at fine-grained level,
we propose to train the entity grounding module in a weakly supervised learning manner.
This can help our system achieve better generalizability to other novel data domains which may just require fine-tuning over dataset annotated coarsely at image level.
This weakly supervised learning can be expressed as selecting the best region $r$ in an image $I$ given an object of interest represented by a textual feature $t$, e.g., a word2vec feature.
With well pre-trained feature extractor, we first extract visual feature maps $v$ over the image, 
based on which we train an attention branch $F$ that outputs
a heatmap expected to highlight a matched region in the image. 

Mathematically, we are interested in obtaining the region $R = F(t, v)$ in the format of heatmap and making sense of it. In practice, we find training a classification model at image level with the attention mechanism works well for entity grounding, which is the output through the attention maps, as illustrated by Figure~\ref{fig:pipeline} left. Moreover, rather than using a multiplicative gating layer to make use of the attention map, we find that it works better by using a bilinear pooling layer \cite{lin2015bilinear,gao2016compact,kong2017low}.

For bilinear pooling, we adopt the Multimodal Compact Bilinear (MCB) pooling introduced in~\cite{fukui2016multimodal} that effectively pools over visual and textual features. In MCB, the Count Sketch projection function \cite{charikar2002finding} $\Psi$ is applied on the outer product of the visual feature $v$ and an array repeating the word feature $t$ for dimensionality reduction: $\Psi (t)\ast \Psi(v)$. If converted to frequency domain, the concatenated outer product can be written as: $\Phi= {FFT}^{-1}(FFT(\Psi (t))\odot FFT(\Psi(v)))$. Based on $\Phi$, the final 2D attentive map $R$ is computed through several nonlinear 1$\times$1 convolutional layers : $R$ = $conv(\Psi)$, with the final one as sigmoid function to shrink all values into $[0,1]$. Later we retrieve the regional representation $f$ by a global pooling over the element wise product between entity attentive map and original visual feature maps: $f = pool(R\odot v)$, on which the weakly supervised classification loss is applied.
Overall, to train the entity grounding module with the attention mechanism in a weakly supervised learning fashion,
we train for image-level $K$-way classification using a cross-entropy loss.

\subsection{Semantic Attribute Grounding Module (M$_{a}$)}
The semantic attribute grounding module improves interpretability of the whole textual
grounding system by explaining that it explains how the final decision is being made. For example, a model finding the ``man in black suits'' as shown in Figure~\ref{fig:ca_demo} should not only output the final grounding mask, but also explain how the final result is being achieved by showing where ``man'' and ``black suits'' are localized in the image.

We also train this module with a weakly supervised learning protocol with similar architecture in the entity module.
But instead of training with $K$-way classification over $K$ predefined attributes as in training entity grounding module, we model this as a multi-label problem, since an image may deliver multiple attributes which are not exclusive to each other. Moreover, rather than classifying them, we propose to use regression for training, since attributes can become large in number while the features representing attribute names can lie in a manifold in the semantic space. This makes our module extensible to more novel attributes even trained with some pre-defined ones.

Note that we represent each attribute with the word2vec feature \cite{mikolov2013distributed}.
Although the word2vec model demonstrates very semantic grouping on words,
we find that these features representing attributes do not deliver reasonable discriminativeness. For example, in word2vec features,
``man'' is more similar to ``woman'' than ``boy'' but we care more about the gender meaning in practice.
Though retraining such a word2vec model solves the problem,
we adopt an alternative method in this paper by proposing a dictionary based scoring function
over the original word2vec features.
We note that this method not only offers more discriminative scoring power,
but also inherits the semantic manifolds in word2vec features,
extensible to novel attributes without re-training whole model as done in $K$-way classification.

To introduce our dictionary based scoring function,
we revisit the classic logistic normalization widely used in binary classification as below:

\begin{equation}
	y_i = \frac{1}{1+\exp(-\w_i^T\x)}
\end{equation}

where ${\bf w}_i$ here represents the learning parameters, 
and ${\bf x}, y_i$ are the input vectors and predicted probability with respect to class $i$.
Note again that, although the logistic loss works well for binary classification
or multi-label classification,
it is not extensible to novel classes unless retraining the whole model.
Our solution to this is based on the proposed dictionary based scoring function.
Suppose there are $C$ attributes, represented by word2vec and stacked as a dictionary
${\bf D}=[{\bf d}_1, \dots, {\bf d}_C]$.
We can measure the (inverse) Euclidean distance between $\x$ and each dictionary atom
for the similarity about which attribute $\x$ is predicted.

So the dictionary acts as the parameter bank
which can be fixed if we want to preserve the
semantic manifold in the word2vec feature space,
and we have the following modified sigmoid transformation:
\begin{equation}
	y_i = \frac{2}{1+\exp(\Vert{\bf d}_i-\x\Vert_2^2)}	
\end{equation}
However, as this may also be less discriminative,
we opt to learn a new latent space.
Concretely, we build new layers before the sigmoid transformation,
and these layers form new function $\phi$ and $\psi$
to transform the feature $\x$ and dictionary atoms, respectively.
Then we have the following dictionary based scoring function for the $i^{th}$ attribute:
\begin{equation}
	y_i = \frac{2}{1+\exp(\Vert \psi({\bf D})_i - \phi(\x)\Vert_2^2)}	
\end{equation}

Furthermore,
despite using the dictionary based scoring function as a modified sigmoid for logistic loss over
the holistic feature globally pooled over the image,
we also perform it at pixel levels.     
Concretely,
during each iteration on each training image,
we choose the $T$ pixels with the top scores to feed into the logistic loss.
This practice is essentially a multi-instance learning at pixel level ~\cite{paul2018w}.
We find in our experiment that jointly using the two losses helps generate
better attention maps.

\subsection{Color Grounding Module (M$_{c}$)}
When querying in natural languages,
human beings typically rely on textual descriptions for low-level vision characteristics,
e.g., color, texture, shape and locations.
Recent work also demonstrates the feasibility of grounding low-level features in unsupervised learning~\cite{vondrick2018tracking}.
In our work for the datasets we studied in our work,
we notice that color is the most used one.
In the Flickr30k Entities dataset~\cite{plummer2015flickr30k} as studied in this paper, 
70\% attributes words are colors describing persons. 
Therefore, 
without loss of generalization, 
we propose to build a separate color grounding module to
improve the interpretability of the whole textual grounding system.

Different from entity grounding and semantic attribute grounding modules,
we train this color grounding module in a fully supervised way over a small-scale dataset, called Color Name Dataset~\cite{van2007learning},
which contains 400 images with color name annotations at pixel level.
We essentially perform pixel-level color segmentation over the input image
to ground color reference. Moreover, we build this color grounding module over a ResNet50 model~\cite{he2016deep} pretrained on ImageNet dataset~\cite{deng2009imagenet}, and concatenate intermediate features at lower levels for pixel-level color segmentation. We find this works better than combining high-level features. We conjecture the reason is due to that color is a very low-level cue that does not require deep architectures and high-level feature abstraction. This is consistent with what reported in \cite{larsson2016learning}.

\begin{figure*}[t]
\begin{center}
\includegraphics[width=1.0\linewidth]{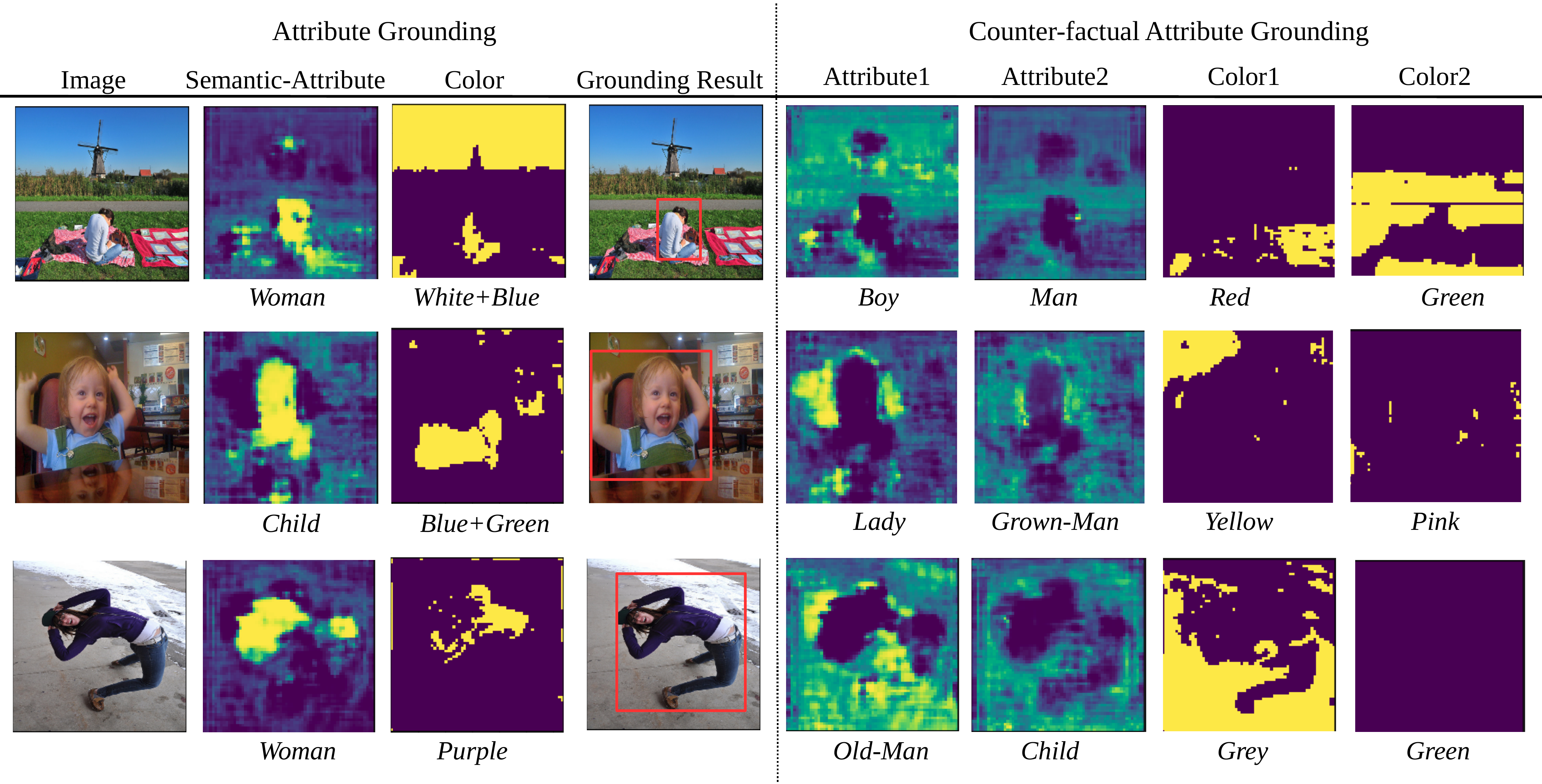}
\end{center}
\vspace{-6mm}
\caption{Examples of attribute grounding predictions (left) and counterfactual attribute grounding results (right). (best viewed in color) }
\vspace{-2mm}
\label{fig:demo}
\end{figure*}

\subsection{Architecture and Training}
Our three modules are based on the ResNet architecture~\cite{he2016deep}.
Similar to~\cite{chen2018deeplab,kong2017recurrent}, we increase the output resolution of ResNet
by removing the top global $7\times 7$ pooling layer and the last two $2\times2$
pooling layers, replacing them with atrous convolution with dilation rate 2 and
4, respectively to maintain a spatial sampling rate.
Our model thus outputs
predictions at $1/8$ the input resolution which are upsampled for benchmarking.
For (multi-label or $K$-way) classification,
we use a global pooling layer that produces a holistic image feature for classification.
In addition, we also insert an $L_{2}$ regularization over the attention maps,
and we observe that such a regularization term helps reduce noises effectively.

We use the standard stochastic gradient decent (SGD) for training in a stagewise fashion.
Specifically,
we first train a plain classification model for entity and semantic attribute grounding modules,
then we build the attention branch for attentional learning.

Though our textual grounding system is end-to-end trainable,
we train each module separately.
And though joint training is straightforward to implement,
we do not do this for practical reasons:
1) we can easily plug in a better trained module without retraining the whole system
for better comparison; 2) we focus on the modular design,
isolating the influence of the settings and parameters of each module.

\begin{figure}[h]
\begin{center}
\includegraphics[width=1\linewidth]{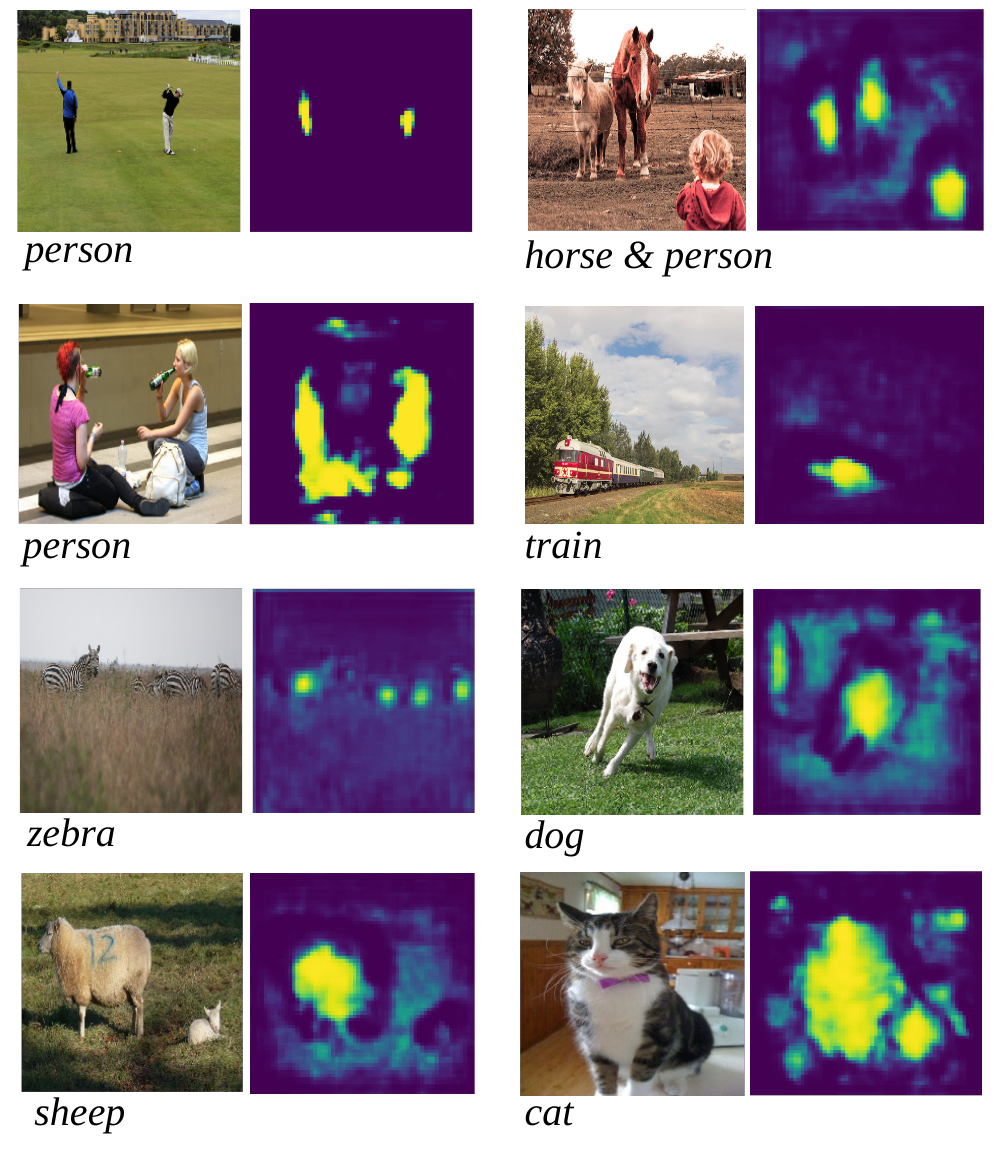}
\end{center}
\vspace{-7mm}
\caption{Qualitative examples of attention maps from the entity module.  }
\vspace{-5mm}
\label{fig:entity_demo}
\end{figure}

\section{Experiments}
\label{sec:exp}
We now experimentally validate our system and compare it with the state-of-the-art methods.
To highlight the generalizability of our system,
we train it on COCO2017 dataset~\cite{lin2014microsoft}
while test it on another Flickr30K Entities dataset~\cite{plummer2015flickr30k}.
We first introduce the two datasets briefly before conducting thorough comparisons,
then we carry out another experiment to show our (weakly supervised) model performs remarkably better than other (fully supervised) methods on a collected dataset consisting of counterfactual testing cases.

\subsection{Datasets and Preprocessing}
The two datasets we used in our experiments are:
COCO2017~\cite{lin2014microsoft} for training our system
and Flickr30k Entities Dataset \cite{plummer2015flickr30k} for testing it.

COCO2017 dataset contains 110k training images with 80 object categories at image level.
These 80 object categories are used for training our entity grounding module as they can be
seen exclusive to each other.
The captioning task and the annotations provided in COCO2017 enables us to train
our semantic attribute grounding module. Using \cite{bird2009natural,miller1998wordnet}, we tokenize and mine out words related to semantic attributes
(e.g., man, woman, boy, old and young) to form our corpus.
To train the semantic attribute grounding module,
we retrieve images from COCO2017 whose captions contain the attributes existing in our corpus.
Eventually,
10,000 images and 34 attributes are collected from COCO2017 for weakly supervised training our modules.
To alleviate imbalanced distribution of these attributes,
we adopt inverse frequency reweighting during training.

The Flickr30k Entities dataset contains over 31k images with 275k bounding boxes with natural languages descriptions,
and we only use this dataset for testing our system with the bounding boxes.

To carry out counterfactual testing experiment,
we collect a new testing set with images from Flickr30k and RefCOCO+ \cite{kazemzadeh2014referitgame}.
The images only contain persons and relevant attributes (e.g., gender, age, etc),
so we call this dataset Person Attribute Counterfactual Grounding dataset (PACG).
By developing an easy-to-use interface,
we are able to generate counterfactual captions for a given image with the good captions provided by the original dataset. Similar to work in \cite{hendricks2018generating}, we generate counterfactual attributes by mining the negation of existing attributes. The overall PACG dataset consists 2,000 images,
a half of which are with counterfactual attributes not existing in the image
and the other half with ``correct'' attributes.

\noindent{\bf Language Processing:}
To deal with free-form textual queries,
we use a language parser ~\cite{bird2009natural} to select the keywords according to the functionalities of the three modules.
We first extract the entity words and pick the most similar object classes by word similarities.
We then extract the semantic attribute words in the same way.
Finally, we extract the the color keywords simply for the color grounding.
To represent the textual attributes and color names,
we adopt the word vectors from GloVe \cite{pennington2014glove}.
This enables meaningful similarity between the defined attributes/colors and
novel ones when encountered at testing stage.

\newcommand\Tstrut{\rule{0pt}{2.2ex}}         
\newcommand\Bstrut{\rule[1.6ex]{0pt}{0pt}}   

\begin{table}[htbp]
\begin{center}
\begin{tabular}{c c c}
\thickhline
 Aprroach & Image Features & mAP (\%)\\
\hline
\vspace{1mm}
\textbf{Supervised} \Tstrut\\
SCRC \cite{hu2016natural} & VGG-cls & 27.80\\
GroundeR$_{s}$ \cite{rohrbach2016grounding} & VGG-cls & 47.81\\
CCA \cite{plummer2015flickr30k} &  VGG-det & 50.89\\
IGOP \cite{yeh2017interpretable} &  YOLO+DeepLab & \textbf{53.97}\\
\hline
\textbf{Unsupervised} \Tstrut\\
Largest proposal & n/a &  24.34\\
GroundeR$_{u}$ \cite{rohrbach2016grounding} & VGG-det & 28.94\\
Mutual Info. \cite{zitnick2013learning} & VGG-det & 31.19\\
UTG \cite{yeh2018unsupervised} & VGG-det & 35.90\\
UTG \cite{yeh2018unsupervised} & YOLO-det & 36.93\\
\hline
\textbf{Weakly-Supervised}\Tstrut\\
Ours$^{1}$  & Res101&  29.01\Tstrut\\
Ours(Attr)   & Res101& 32.04\\
Ours(Attr+Col)  & Res101& 33.43\\
Faster-RCNN\cite{ren2015faster} & Res101-det& 35.35\\
Ours+Attr   & Res101-det& 47.46\\
Ours+Attr+Col  & Res101-det& \textbf{48.66}\\
\thickhline
\end{tabular}
\end{center}
\vspace{-5mm}
\caption{Phrase localization performance on Flickr 30k Entities (accuracy
in \%). }
\label{table:flickr}
\end{table}

\subsection{Textual Grounding Evaluation}
We compare our modular textual grounding system with
other supervised/unsupervised methods on the Flickr30k Entities dataset.
We use the mean average precision (mAP) metric to measure the quantitative performance.
The detailed comparison is listed in Table~\ref{table:flickr}.

As the first baseline method similar to~\cite{yeh2018unsupervised},
we select the largest proposal as the final result.
This method achieves 24.34\% mAP.
Then, we build another baseline model that
we train the entity grounding module only through weakly supervised learning over a ResNet101 backbone, which is pretrained over ImageNet dataset. Then, over the entity grounding heatmaps,
we generate bounding boxes candidates by sub-window search \cite{lampert2009efficient} together with contour detection results, followed by a Non-Maximum Suppression to further refine the proposal boxes. We select the box that encompasses largest ratio of object according to equation \ref{eq:att}. We note that this simple baseline module (29.01\% mAP) outperforms GroundR$_{u}$~\cite{rohrbach2016grounding} (28.94\% mAP) that learns grounding in an attentive way over large-scale training data.
If we include our semantic attribute module,
we improve the performance further (32.04\% mPA),
outperforming Mutual Info.~\cite{zitnick2013learning}. If we further insert the color grounding module,
we achieve comparable performance (33.43\%) to UTG (36.93\% mAP),
which adopts an unsupervised method to link image concepts to query words~\cite{yeh2018unsupervised}.
We note that our models are trained on COCO dataset only,
unlike all these methods which are trained on the same dataset (Flickr30k dataset).
The effectiveness of our model is demonstrated by its good transferability, as it is trained and tested
on different data domains.

It is also worth noting that, all the compared unsupervised methods unanimously adopt a well-trained object detector, even though they claim to be unsupervised learning.
To gain an idea how the detector improves the performance,
we fine-tune the faster-RCNN detector~\cite{girshick2015fast} on COCO
and train our modules with weak supervision again.
We report our results as the bottom two rows in Table \ref{table:flickr}.
Now we can see our models perform significantly better, and even surpasses some fully supervised methods (SCRC~\cite{hu2016natural}
and GroundeR~\cite{rohrbach2016grounding}).
Although it seems unfair that our system adopts ResNet101 architecture
while most compared methods uses shallower VGG networks,
we note that IGOP which adopts both VGG and ResNet101 (denoted by DeepLab) 
achieves the best performance with fully supervised training.
Even though our best model does not outperform IGOP,
we believe the performance gap is small and reasonable as our training is carried out
on a different dataset (COCO) rather than Flickr30k, and it does not
rely on any strong supervision signals. We show output examples of entity grounding module in Figure \ref{fig:entity_demo} with various object categories as input, and attribute grounding outputs in Figure \ref{fig:demo}, with both existing attributes and counterfactual attributes as queries. These visualizations demonstrates how our system rejects  in an explainable way the counterfactual queries through the modular output.

\subsection{Counterfactual Grounding Evaluation}
\begin{figure}[t]
\begin{center}
\includegraphics[width=1.1\linewidth]{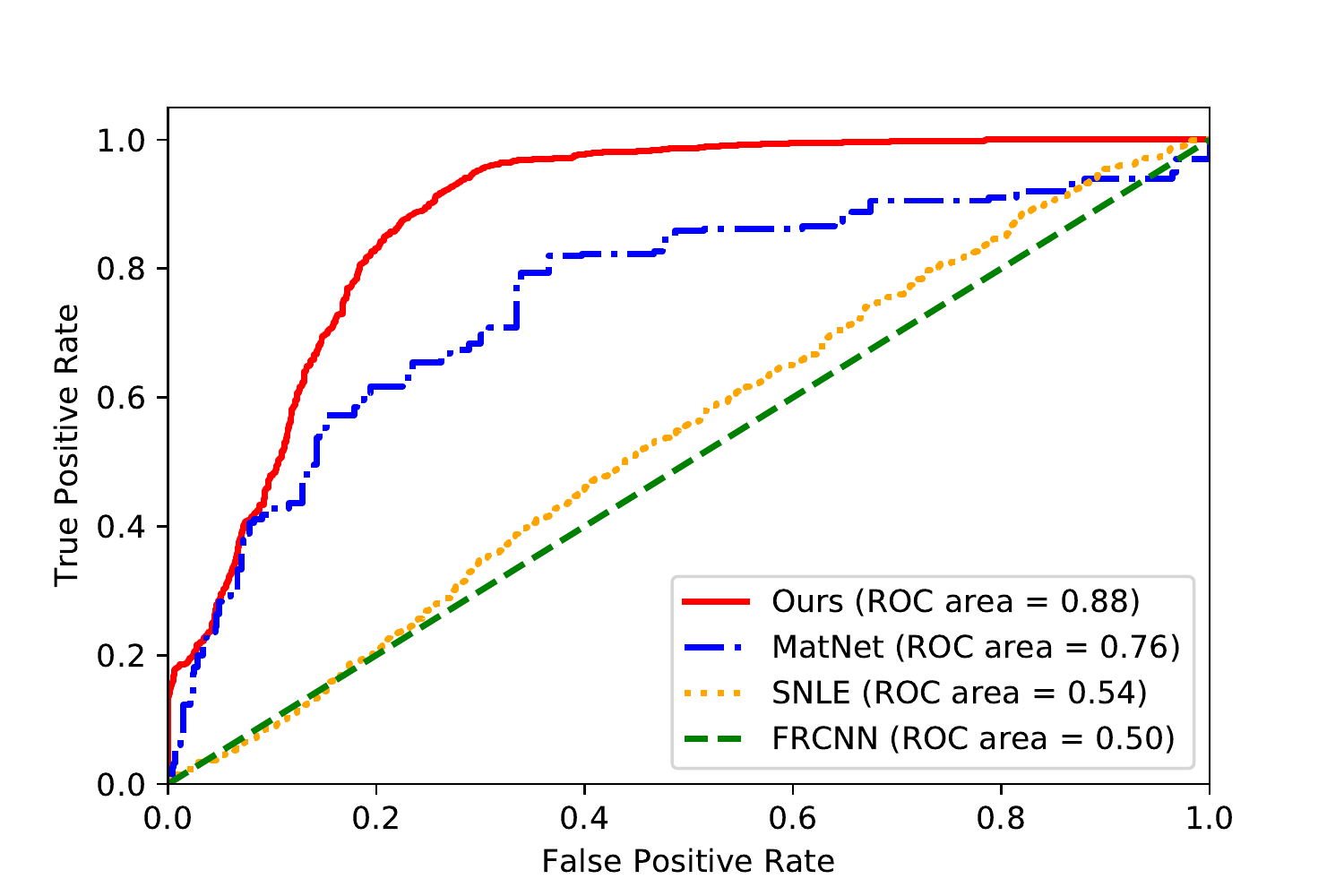}
\end{center}
\vspace{-6mm}
\caption{ROC of our modular network demonstrates high resolving ability on PACG dataset with an AUC of 0.88, comparing to other state of the art baseline models (best viewed in color).  }
\vspace{-2mm}
\label{fig:roc}
\end{figure}

We now carry out in-depth study on how our system performs when facing of counterfactual textual queries over our collected
PACG dataset, and compare with three baseline or state-of-the-art methods, Faster-RCNN~\cite{ren2015faster}, MattNet \cite{yu2018mattnet}, SNLE \cite{hu2016segmentation}. We plot the ROC curves for these methods in Figure~\ref{fig:roc}. Textual grounding system then selects the region with highest scores/probability. We compare the prediction scores/probabilities of the predicted regions between the counterfactual queries and normal queries and expecting to observe distinct difference between their numerical scores.

We clearly see from the figure that our system achieves the highest AUC among of these methods, meaning that modular design successfully increases the counterfactual resilience of the grounding system. Specifically, end-to-end models like SNLE \cite{hu2016segmentation} encode the textual query
into a vector representation to extract spatial feature maps from the image as response map. However, such encoding do not consider the internal structure of sentences \cite{macwhinney1997second}, also neglecting semantic nuances of near-synonyms. Note that MattNet~\cite{yu2018mattnet} also adopts a modular design, but it is trained with fully supervised learning, also it is not easily extended to novel attributes and unable to reject counterfactual queries as effectively as our method.
The AUC of Faster-RCNN is approximately 0.5 since the recognition ability is restricted to entity-level and not been able to discern among semantic attributes.
We conclude that with the modular design and better scoring function in each modules, our model demonstrated highly resilient ability against counterfactual queries, even with only weakly-supervised training.
\vspace{-2mm}

\section{Extensive Applications}
\label{sec:app}

\begin{figure}[t]
\begin{center}
\includegraphics[width=1\linewidth]{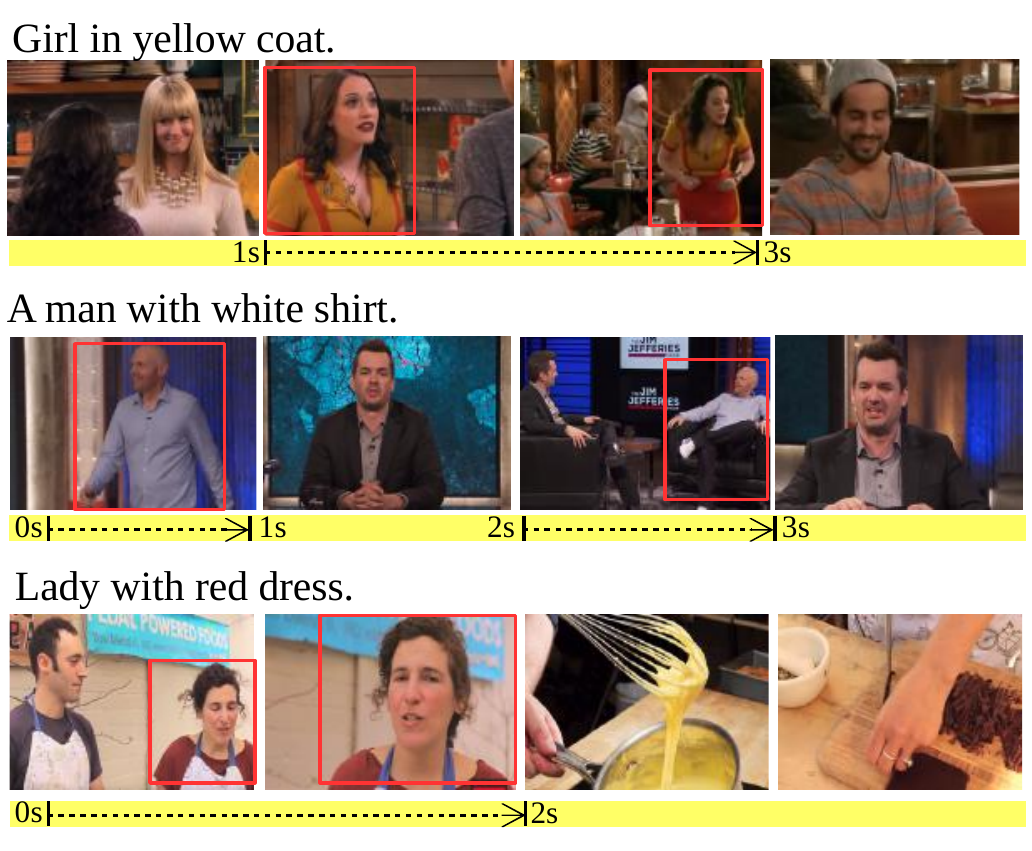}
\end{center}
\vspace{-2mm}
\caption{Temporal/Spatial grounding in video sequences. Time-segments contain phrases are selected to filter out irrelevant frmaes.}
\label{fig:video}
\vspace{-6mm}
\end{figure}

The counterfactual resilient design can be furthered applied to various tasks. 
In this section we showcase some practical applications.

\noindent{\bf Grounding Textual Phrase in Video} 
To ground textual phrase in video, 
the system needs to first determine which temporal segment and moment to 
retrieve \cite{hendricks2017localizing}, 
then localize the region associated with the descriptions. 
In this case, textual information may be irrelevant to most of the video frames, 
thus requiring the system to be counterfactual resilient to query 
and discern whether it is existing or not in the current segment. 
Unlike an existing approach~\cite{gavrilyuk2018actor}, 
which treats the problem as temporal localization, 
we score a set of frames and select out segments that are more likely to be relevant to sentence. 
We demonstrate this process in Figure \ref{fig:video} 
that modular network successfully conduct a temporal-spatial 
grounding task in video clips.

\begin{figure}[t]
\begin{center}
\includegraphics[width=1\linewidth]{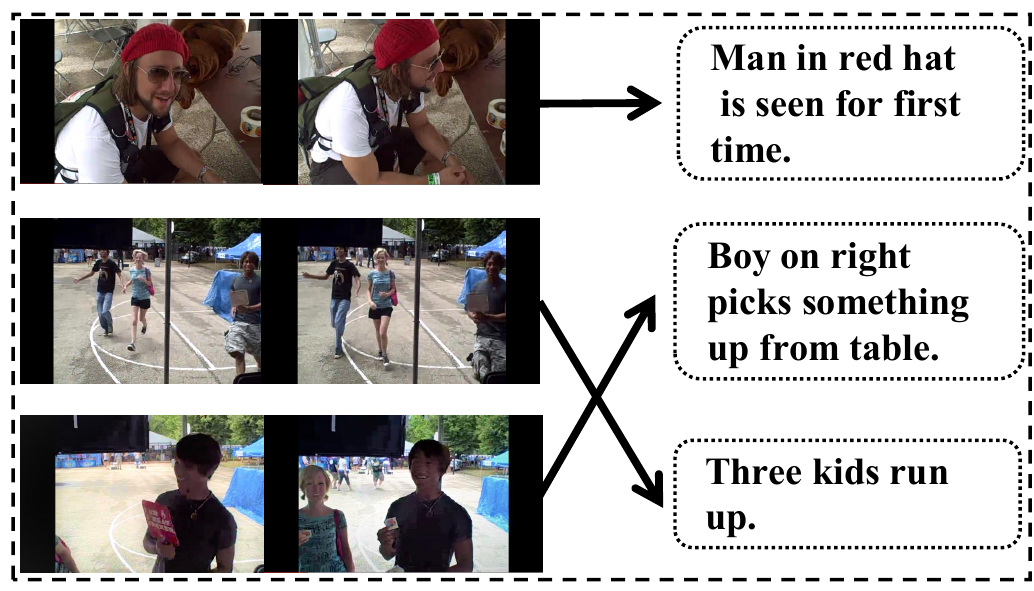}
\end{center}
\vspace{-2mm}
\caption{Video captioning alignment. With unordered captions, our system links each sentence to it's corresponding frames. Examples took from DiDeMo~\cite{hendricks17iccv}. }
\label{fig:relink}
\end{figure}

\noindent{\bf Video to Captioning alignment} Our model can be used to correct misaligned captioning sentences like the work in~\cite{bojanowski2015weakly}. Given mis-matched frames and captions, we examine the sentence-frame relevance and find the corresponding frame for each sentence. Figure \ref{fig:relink} shows an example of the captioning alignment, the temporal linked sentences can be re-ordered based on video.
\vspace{-3mm}

\section{Conclusion }
\label{sec:conclusion}
\vspace{-2mm}

In this paper, we propose to modularize the complex textual grounding system
by decomposing the textual description/query into three parts:
entity, semantic attributes and color.
Such a modular design largely improves the interpretability and counterfactual resilience of the system.
Moreover,
we propose to train the modules in a weakly supervised way,
so we merely needs image-level labels which are easy to obtain.
This largely helps alleviate the requirement of large-scale manual annotated images for training,
and for fine-tuning if transferring the system to a new data domain.
Through extensive experiments,
we show our system not only surpasses all unsupervised textual grounding methods
and some of fully supervised ones,
but also delivers strong resilience when facing counterfactual queries.

Our modularized textual grounding system is of practical significance as it can be deployed
in various problems.
In this paper,
we show how our system can be applied to video captioning correction and visual-textual search.
We expect more applications can benefit from our modular design.
\newpage
\noindent{\bf Acknowledgements: }The support of the National Science Foundation under the Robust Intelligence Program (1816039 and 1750082), and a gift from Verisk AI are gratefully acknowledged. We also acknowledge NVIDIA for the donation of GPUs. 




\begin{center}
 {\large \textbf{Appendix}}
\end{center}

In this appendix, we provide details of the Person Attribute Counterfactual Grounding dataset (PACG), and show more visualizations of  modular groundings and results on counterfactual queries.
\section{PACG} 
\begin{figure*}[h]
\begin{center}
\includegraphics[width=1.0\linewidth]{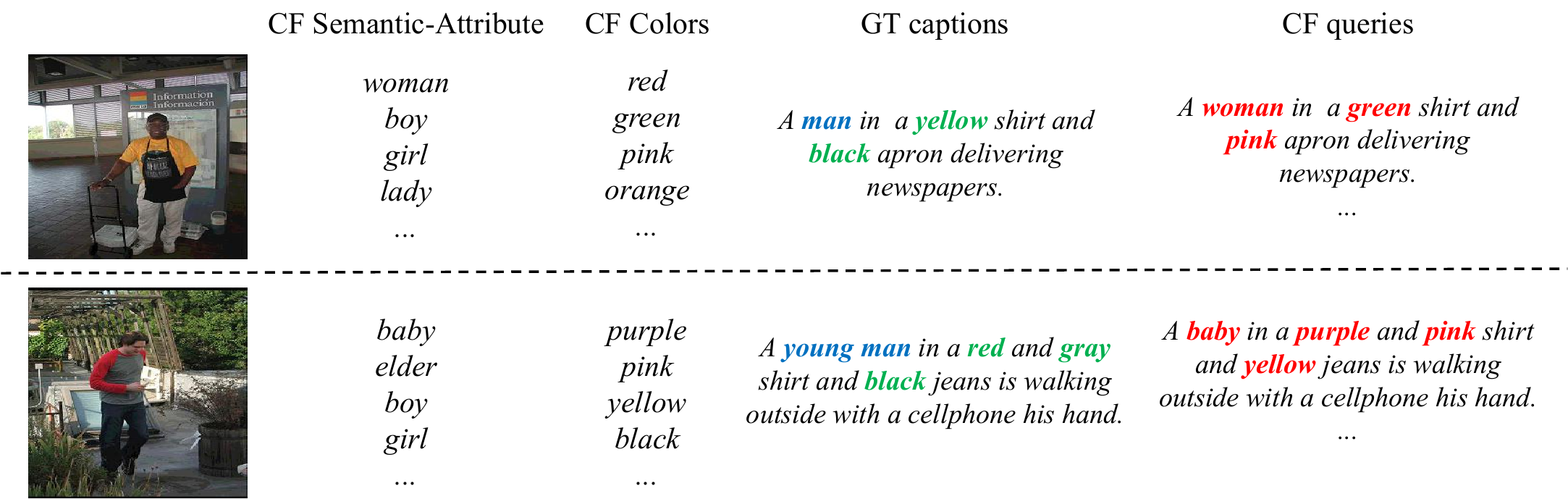}
\end{center}
\caption{Illustrative examples of PACG dataset: we first annotate all existing semantic attributes and colors based on captions and manually checking. Counterfactual (CF) attributes and queries are then generated based on that.}
\vspace{-2mm}
\label{fig:sup1}
\end{figure*}

To collect counterfactual textual grounding test set, we select 2k images from Flickr 30k Entities Dataset~\cite{plummer2015flickr30k} and RefCOCO+ Dataset~\cite{kazemzadeh2014referitgame}. We first extracted all attribute words from the ground truth captions as existing attributes. Whereafter, we further manually check and complement all the missing attributes in the images on an interactive user interface. We demonstrate several examples from PACG in Figure~\ref{fig:sup1}, and their counterfactual attributes/queries. We adopt person-related semantic attributes and colors on them as the grounding queries. All colors and words from our corpus, that are not related to the person showing up the image, are considered as counterfactual (CF) attributes in the specific image (see Figure~\ref{fig:sup1}). To conduct textual grounding evaluations, we substitute the original attribute words in the ground truth captions with CF attributes (semantic-attributes or colors) as our CF queries. We generate all the possible CF queries per image, and more than 20k CF queries are obtained in total.

\begin{figure}[h]
\begin{center}
\includegraphics[width=.99\linewidth]{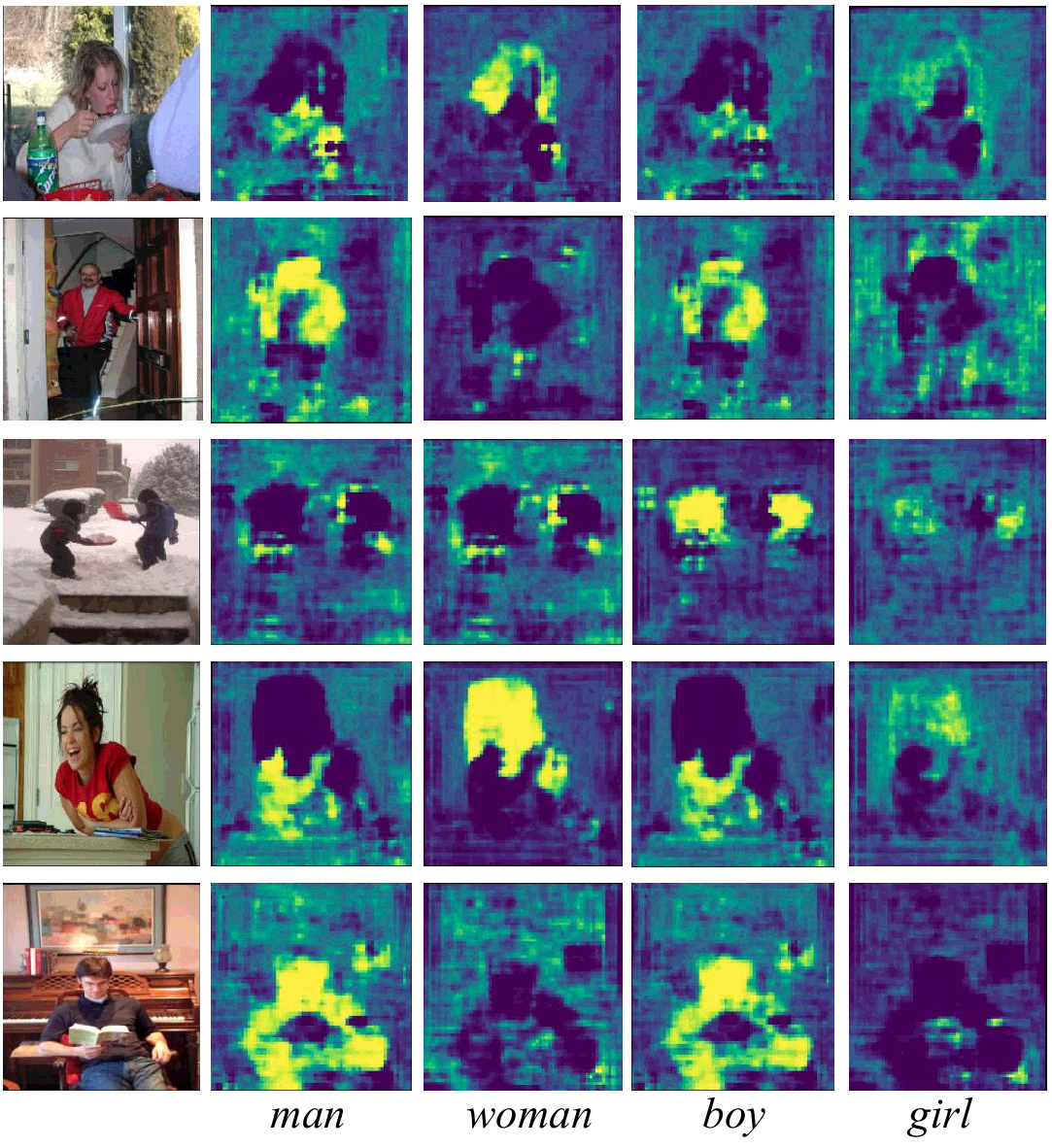}
\end{center}
\vspace{-4mm}
\caption{Examples of our grounding results from the semantic attribute grounding module.}
\vspace{-6mm}
\label{fig:sup3}
\end{figure}

\begin{figure}[h]
\begin{center}
\includegraphics[width=.99\linewidth]{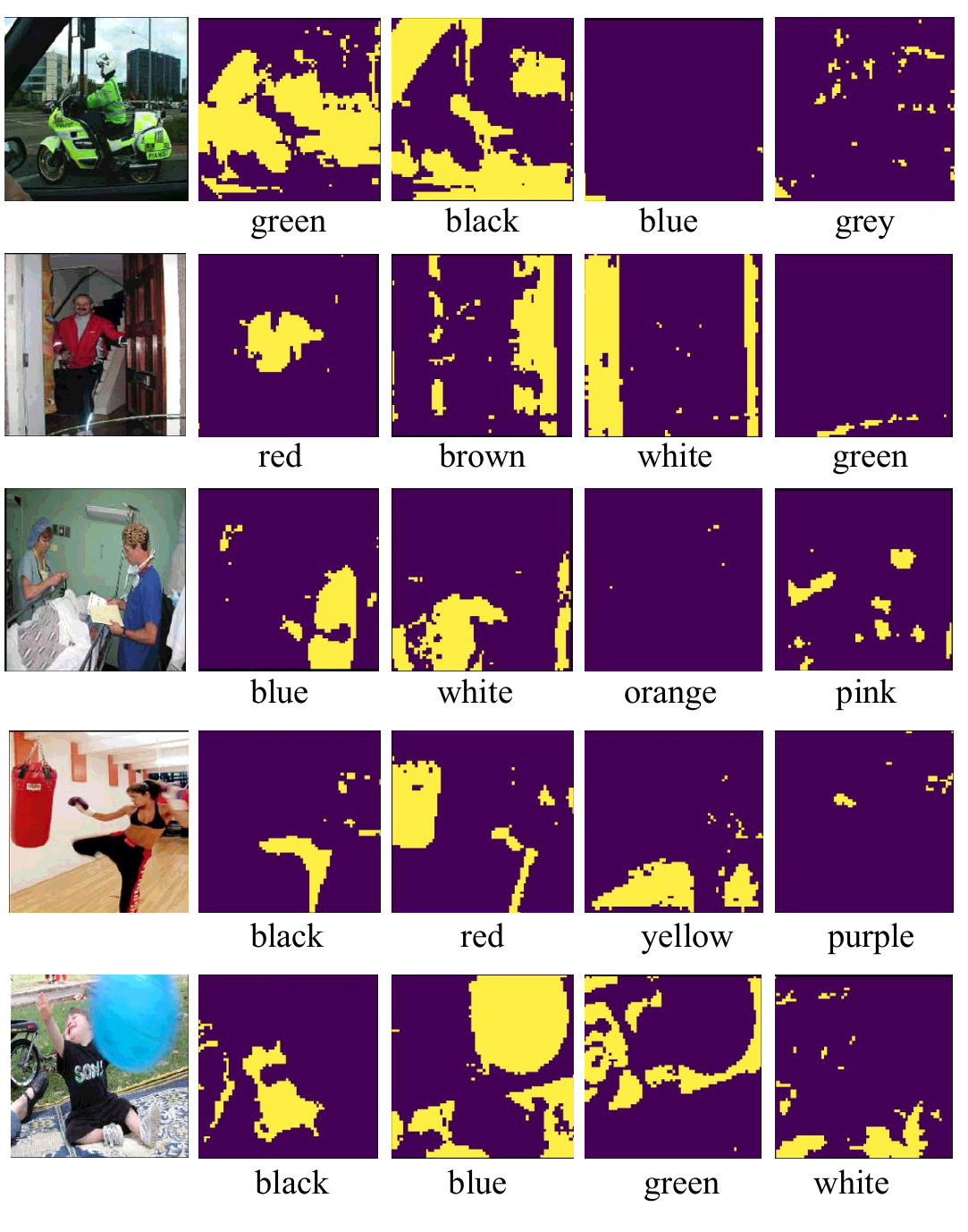}
\end{center}
\vspace{-8mm}
\caption{Examples of our grounding results from color grounding module.}
\label{fig:sup4}
\end{figure}

\begin{figure*}[h]
\begin{center}
\includegraphics[width=1.0\linewidth]{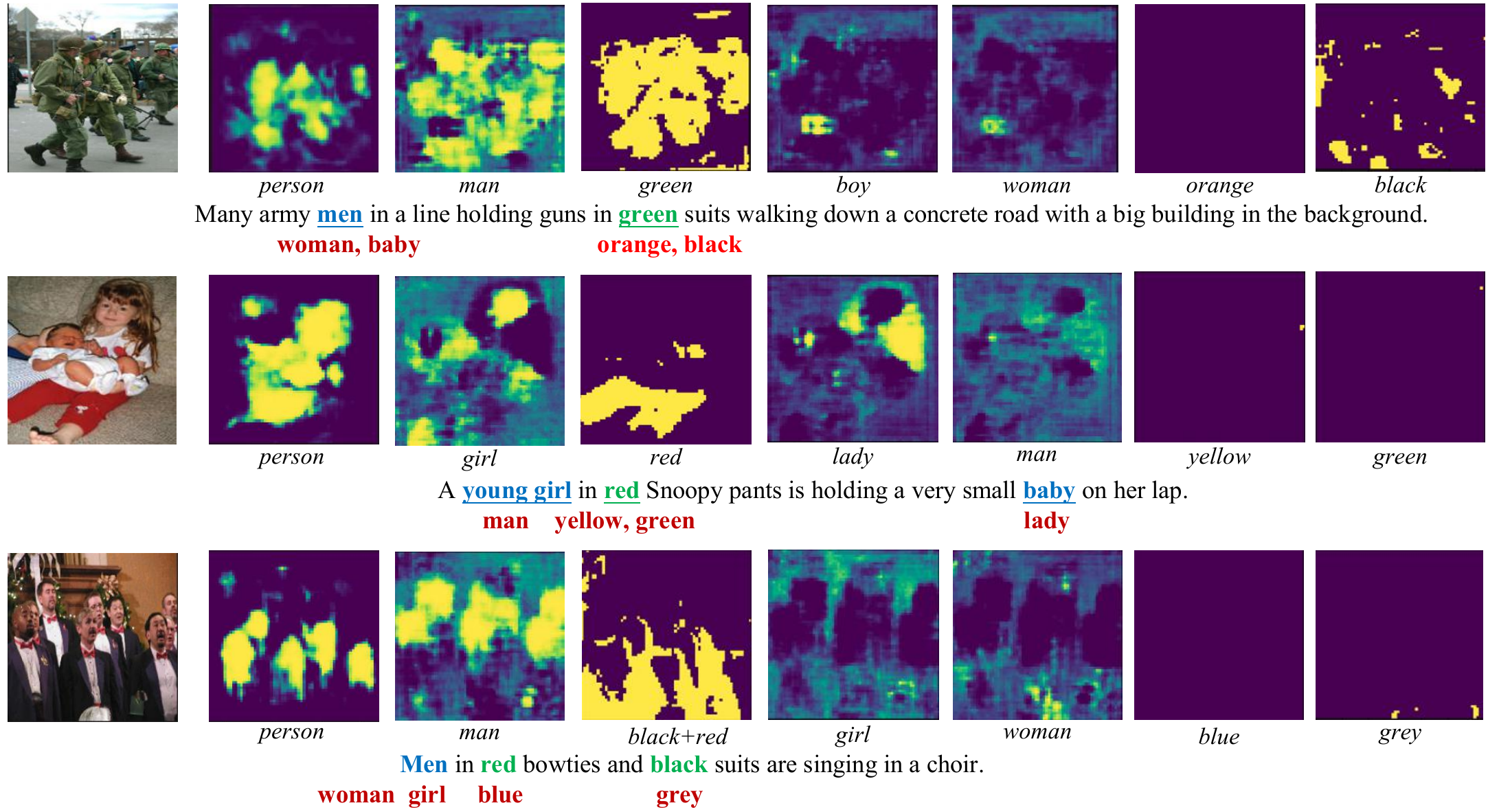}
\end{center}
\vspace{-4mm}
\caption{Examples of our grounding results on counterfactual queries. We replace the original attribute words with our CF attributes from PACG. Starting from leftmost row, we show the original images, outputs from entity grounding module, followed by ground truth attribute grounding results, and CF attribute grounding outputs. }
\vspace{-2mm}
\label{fig:sup2}
\end{figure*}

\section{Additional Grounding Results from Sub-modules}
\vspace{-1mm}
We show additional grounding outputs from our semantic attribute grounding module (M$_a$) and color grounding module (M$_c$) in Figure~\ref{fig:sup3} and Figure~\ref{fig:sup4}. For semantic grounding module, we visualize the attention maps of four most frequent input queries ``man'', ``woman'', ``boy'', and ``girl'' (see Figure~\ref{fig:sup3}). We see clearly from Figure~\ref{fig:sup3} that, our module distinguishes between textual words with different gender and age. It indicates that the semantic attribute grounding module is able to learn the concepts of gender and age through weakly supervised training.

\section{Additional Grounding Results on Counterfactual Queries}
We demonstrate additional intermediate grounding outputs on PACG dataset based on our modularized network in Figure~\ref{fig:sup2}. Specifically, besides showing the grounding outputs on ground truth attributes in the left four rows, we also show CF attribute grounding outputs from sub-modules in the last four rows. From the attention maps, we observe that on these examples (1) our model precisely give out the region of ground truth attributes; (2) our model successfully reject the counterfactual queries in these cases.

\newpage\newpage

\end{document}